\documentclass[letterpaper, 10 pt, conference]{ieeeconf}  
\pdfoutput=1

\PassOptionsToPackage{bookmarks=false}{hyperref}

\usepackage{graphics} 
\usepackage{epsfig} 
\usepackage{amsmath} 
\usepackage{amssymb}  
\usepackage{color}
\usepackage{ulem}
\usepackage{subcaption}
\usepackage{algorithm}
\usepackage{algpseudocode}
\usepackage{url}
\usepackage{hyperref}

\title{\LARGE \bf
Efficient Self-Supervised Data Collection for Offline Robot Learning
}

\author{Shadi Endrawis${^{12}}$, Gal Leibovich${^2}$, Guy Jacob${^2}$, Gal Novik${^2}$, Aviv Tamar${^1}$ \\

$^1$ Technion -- Israel Institute of Technology\\
$^2$ Intel Labs \\
}

\begin{document}

\maketitle
\thispagestyle{empty}
\pagestyle{empty}

\begin{abstract}

A practical approach to robot reinforcement learning is to first collect a large batch of real or simulated robot interaction data, using some data collection policy, and then learn from this data to perform various tasks, using offline learning algorithms. Previous work focused on manually designing the data collection policy, and on tasks where suitable policies can easily be designed, such as random picking policies for collecting data about object grasping. For more complex tasks, however, it may be difficult to find a data collection policy that explores the environment effectively, and produces data that is diverse enough for the downstream task. 
In this work, we propose that data collection policies should actively explore the environment to collect diverse data. In particular, we develop a simple-yet-effective goal-conditioned reinforcement-learning method that actively focuses data collection on novel observations, thereby collecting a diverse data-set. We evaluate our method on simulated robot manipulation tasks with visual inputs and show that the improved diversity of active data collection leads to significant improvements in the downstream learning tasks.
\end{abstract}

\section{INTRODUCTION}\label{s:intro}
Reinforcement learning (RL) is a popular approach for learning robotics skills \cite{kober2013reinforcement, levine2016end, qt-opt}. In general, RL can be performed in an \emph{online} manner, where the robot simultaneously interacts with the environment and improves its policy, or \emph{offline}, where first, some data about the robot-environment interaction is collected, and then, a learning algorithm calculates the desired policy using this data.

The offline learning approach offers many practical benefits: data can be collected in a safe and controlled manner, and collection can be parallelized across different robots/environments~\cite{lange2012batch,levine2020offline}. However, the quality of offline learning clearly depends on the quality of the data, which in turn depends on the data collection policy. Indeed, recent studies focused on tasks where a policy that produces high-quality and diverse data can manually be designed, such as randomly picking objects for learning to grasp~\cite{qt-opt}, randomly poking objects for learning to push~\cite{agrawal2016learning,wang2019learning,finn2017deep}, or randomly throwing an object for learning to grasp and throw~\cite{zeng2020tossingbot}.

For learning more complex tasks, however, we posit that manually designing data collection policies can become very difficult. As an example, consider a robot that must learn to stack two rigid blocks. If data is collected by a policy that randomly picks and places a block, the chances of observing task-relevant data is small. In this work, we investigate a principled approach for designing data collection policies that actively explore their environment, with the hypothesis that active exploration can lead to more relevant data.

We aim for a general method that can produce high-quality data-sets that would broadly improve the performance of downstream robot learning tasks.
Our specific desiderata are:
\begin{enumerate}
    \item Data that exhibits a diversified set of behaviours to allow for building better models and policies.
    \item Data that is evenly distributed, and not highly concentrated on a small part of the state space.
    \item Efficient data collection, without spending valuable robot time on non-interesting parts of the state space.
\end{enumerate}

We approach the problem by defining the data collection phase as a reinforcement learning (RL) problem, where the goal of the agent is to cover as much of the state space as possible. Such objectives are common in RL exploration strategies based on intrinsic motivation (IM)
\cite{conf/nips/BellemareSOSSM16, 10.3389/neuro.12.006.2007, DBLP:books/sp/13/Barto13, DBLP:conf/icml/PathakAED17, curiosity-driven, rnd}.
We propose an improvement of the random network distillation (RND) method~\cite{rnd} that actively drives the robot to explore novel states, thereby efficiently collecting diverse data. Our key idea is that we can use goal-conditioned policies to quickly reach states that are novel, without waiting for the robot to reach these states through the slow reward maximization of standard RL.

\begin{figure}
    \centering
    \includegraphics[width=0.95\columnwidth]{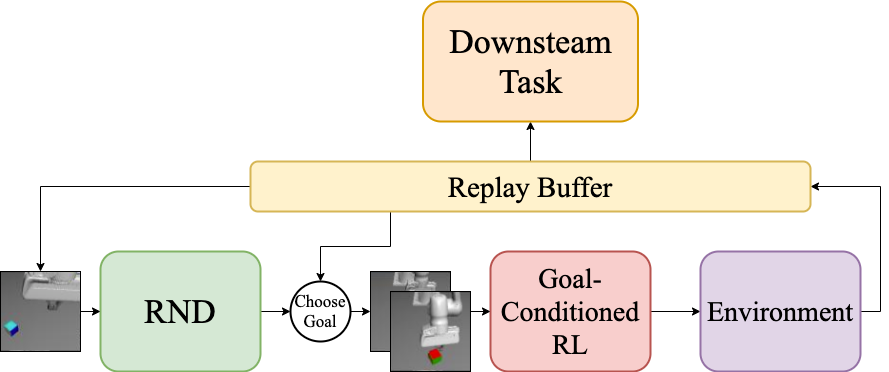}
    \caption{Goal-Conditioned Exploration for Data Collection: We show that we can improve the learning of downstream tasks by using higher quality data that we produce using an exploration scheme that combines intrinsic motivation and goal-conditioned policies.}
    \label{fig:intro_figure}
\end{figure}

We focus on learning from high dimensional image inputs -- a challenging and important robot learning setting. Our empirical results demonstrate significantly better exploration than commonly used RL exploration strategies. More importantly, our data-set collection method significantly improves learning performance in various downstream tasks, such as supervised learning and offline RL.

\section{BACKGROUND}

We recapitulate several RL ideas we build on in our work:
\paragraph{Reinforcement Learning} \label{background:rl}
In Reinforcement Learning, at every time step $t$ an agent is in a state $s_t \in  \mathcal{S}$. The agent executes an action $a_t \in  \mathcal{A}$, transitioning into a new state following the transition probability $s_{t+1} \sim p(s_{t+1}|s_t, a_t)$, and receiving a reward $r_t = R(s_t, a_t)$. This is often formalized as a Markov Decision Process (MDP). The goal is to find a policy $a_t = \pi(s_t)$ that maximizes expected cumulative reward, $\mathbb{E}\left[\sum^T_{t=1} \gamma^t r_t \right]$, where $\gamma\in(0,1]$ is the discount factor.
The value of a state-action pair when following a policy $\pi$, $Q^{\pi}(s,a)$,
is defined as the expected sum of discounted rewards
from that state-action pair:
$$
Q^{\pi}(s,a) = \mathop{\mathbb{E}}_{a \sim \pi} \left[ \sum^{\infty}_{k=t+1} \gamma^{k-t-1}r_k \Big\vert s_t=s, a_t=a \right].
$$ 

Actor-critic algorithms aim to find the optimal policy $\pi^*(s_t)$ by using two functions approximators, one for the policy $\pi_{\theta}(s_t)$ (actor) and the other for state-action value function $Q^{\pi}_{\phi}(s_t,a_t)$ (critic). Specifically, we use \textit{Twin Delayed Deep Deterministic Policy Gradient} (TD3) \cite{DBLP:journals/corr/abs-1802-09477}, which learns $Q^{\pi}_{\phi}(s_t,a_t)$ by minimizing the mean squared Bellman error of transition sampled from a replay buffer $\mathcal{D}$:
\begin{equation*}
  \resizebox{0.48\textwidth}{!}{$\mathcal{L}(\phi)=\underset{(s_t,a_t,r_t,s_{t+1}) \sim \mathcal{D}}{\mathbb{E}} \left[\left(r_t +\gamma Q_{\phi}(s_{t+1},\pi_{\theta}(s_{t+1})) - Q_{\phi}(s_t,a_t)\right)^2\right]$} ,
\end{equation*}
and learns a policy $\pi_{\theta}(s_t)$ that maximizes $Q_{\phi}(s_t,a_t)$ using a loss function
$
\mathcal{L}(\theta)=-\underset{s_t \sim \mathcal{D}}{\mathbb{E}} \left[Q_{\phi}(s_t,\pi_{\theta}(s_t))\right].
$
TD3 additionally applies stability improvements such as a target network, twin critics, delayed updates, and noise regularization. 
We use TD3 for data collection, by training policies that maximize intrinsic motivation (Sec.~\ref{ssec:IM}) or goal-conditioned policies (Sec.~\ref{ssec:GB}).

\subsection{Intrinsic Motivation} \label{ssec:IM}
Intrinsic motivation (IM) methods add a reward signal to an RL agent that incentivizes exploratory behavior. Curiosity, one of the most widely used IM methods, adds reward 
for visiting novel states. For high dimensional problems, the `novelty' of the state can be hard to define, and several ideas have been explored in the literature \cite{conf/nips/BellemareSOSSM16,DBLP:conf/icml/PathakAED17,DBLP:journals/corr/OstrovskiBOM17}; here we focus on random network distillation (RND)~\cite{rnd, curiosity-driven}. The RND reward bonus is the error in predicting the features of the current state, where the features are defined by a randomly initialized `target' neural network that is not changed throughout learning. A `predictor' network with the same architecture as the target network is trained to predict the target output on data collected by the agent. Formally, the target network maps a state to an embedding $f:\mathcal{S} \rightarrow \mathbb{R}^k$ and the predictor neural network $\hat{f}:\mathcal{S} \rightarrow \mathbb{R}^k$ is trained by gradient descent to minimize the expected MSE loss $\Vert\hat{f}(s_t\vert\theta_{\hat{f}}) - f(s_t)\Vert^2$ with respect to its parameters $\theta_{\hat{f}}$. The prediction error is expected to be higher for novel states than for states similar to the ones the predictor has trained on.

\subsection{Goal-Conditioned Learning} \label{ssec:GB}
To handle sparse reward RL problems, several recent works proposed learning goal-conditioned polices. The idea is that learning to reach \emph{any} state in the domain will give a denser reward signal, which can be easier to learn from.
Universal Value Function Approximators (UVFA) \cite{UFA} simply augment the state input to the policy and value function, $s \in \mathcal{S}$, with an additional goal-state input $g \in \mathcal{G}$.
Hindsight experience replay (HER) \cite{HER} is an off-policy algorithm for training goal-conditioned policies. After collecting a trajectory $s_0,s_1,...,s_T$, we store in the replay buffer every transition $s_t \rightarrow s_{t+1}$ not only with the original goal used for this episode, but also with a goal randomly taken from the trajectory $g'\in s_0,s_1,...,s_T$. 
The main idea is that even in trajectories where we failed to achieve the goal we sought to reach, we still have successfully collected data for training the UVFA to reach $g'$.
Notice that the goal being pursued influences the agent’s actions but not the environment dynamics and therefore we can replay each trajectory with an arbitrary goal, assuming that we use an off-policy RL algorithm.

\section{METHOD} \label{s:method}

In this section, we describe our RL-based method for automatic data collection. As discussed in Sec.~\ref{s:intro}, we desire data that effectively covers the state space, and IM-based methods are therefore suitable for driving the agent towards parts of the state space that have not yet been visited. However, by simply adding a reward bonus to the RL algorithm, such methods explore the state space inefficiently: typical RL algorithms adapt slowly to changes in the reward function, and the agent will spend a long time visiting already known parts of the state space until its policy changes to visit the novel parts. For online RL, this issue manifests as a slow learning process. In offline RL, however, where the data set size is limited in advance, the problem is more severe: inefficient learning will cover less relevant states, and will therefore hinder performance in downstream tasks.

We propose an algorithm for an agent to autonomously explore its domain and reach novel states quickly. Our main idea is to actively drive the agent towards novel states by training a goal-conditioned policy, and set the goals to be states which are deemed novel according to the current RND criterion. Thus, when a new part of the state space is identified as novel, the agent will not spend time on learning to reach it through maximizing the reward function, but will directly navigate towards it using the goal-conditioned policy.

\begin{algorithm}[t]
\caption{Data-set Collection with Goal-Conditioned Policy and RND}\label{alg1}
\begin{algorithmic}[1]
\small
\State Initialize RND networks $f, \hat{f}$
\State Initialize off-policy algorithm $\mathcal{A}$ with replay buffer $\mathcal{R}$
\For{training cycle $=1,\dots,N$}
    \For{episode $=1,\dots,M$}
    \State Sample $\mathcal{R}^{\prime}$ from $\mathcal{R}$ \Comment{$\vert \mathcal{R}^{\prime} \vert = M\cdot T$}
    \State $g\leftarrow$ argmax$_{\mathcal{R}^{\prime}}\Vert \hat{f}(s_{k+1})-f(s_{k+1})\Vert^2$
        \For{$t=0,\dots,T$}
        \State $a_t\leftarrow \pi(s_t, g)$
        \State Execute the action $a_t$ and observe new state $s_{t+1}$
        \State Store transition $(s_t, a_t, s_{t+1})$ in $\mathcal{R}$
        \EndFor
        \For{$0,\dots,T$}
        \State Sample mini-batch $B$ from $\mathcal{R}$
            \For{$(s_{\tau},a_{\tau},s_{\tau+1})\in B$}
            \State Sample goal $g_{\tau}$ from $s_{\tau},...,s_T$
            \State $r_{\tau}\leftarrow -(g_{\tau} \neq s_{\tau})$
            \State $d_{\tau}\leftarrow (g_{\tau} = s_{\tau})\vee(g_{\tau} = s_{\tau+1})$ \Comment{done flag}
            \State Store $\left((s_{\tau}, g_{\tau}), a_{\tau}, r_{\tau}, (s_{\tau+1}, g_{\tau}), d_{\tau}\right)$ in $B^{\prime}$
            \EndFor
            \State Perform one step of off-policy RL using $\mathcal{A}$ and $B^{\prime}$
        \EndFor
    \EndFor
    \State Train $\hat{f}$ on the last $M$ episodes in $\mathcal{R}$ for $K$ epochs
\EndFor
\State\Return $\mathcal{R}$

\end{algorithmic}
\end{algorithm}

Our method is specified in Algorithm~\ref{alg1}. To act in the environment, at the beginning of every episode the agent picks a goal state with maximal novelty from the replay buffer (line 6), and then uses the goal-conditioned policy to reach that goal. In practice, it is usually infeasible to calculate novelty for the entire replay buffer $\mathcal{R}$ every episode, so in lines 5-6 the maximum is taken only over a subset $\mathcal{R}^{\prime}$.

To train the goal-conditioned policy using an off-policy algorithm, we follow an approach similar to HER, which samples a goal from the future of the state in the episode and assigns -1 reward to any transition that does not reach the goal. This self-supervised reward definition causes the agent to learn to reach its goal with as few steps as possible. Under such definition, the goal-conditioned value function of the off-policy algorithm is effectively an inverse distance function between two state observations. 

However, dissimilar to HER, which samples a goal for each state and stores the state-goal pair statically in the replay buffer (i.e., for each state the goal is chosen once when the state is being inserted into the replay buffer, corresponding to setting the goal before line 10 in Algorithm~\ref{alg1} instead of in line 15), we sample goals dynamically each time a batch of states is sampled to train the network, which means that each state can be paired with many different goals throughout the training. We found that this allows for greater sample efficiency and better use of the data.

Additionally, with a small probability, for state $s_{\tau}$ we sample the goal to be the state itself, and in that case the reward is 0 and the done signal is $True$ (lines 16 and 17 in Algorithm~\ref{alg1}), meaning that the value of the state-goal pair should be 0. We found that this helps stabilize training. We hypothesize that since the value function represents an inverse distance function, if the network never encounters identical or very close state-goal pairs during the training, the predictions near the goal would be very noisy. 

In order to not be biased towards states that remain in the buffer for a long time, RND is trained only on states from recent experience, similar to the online RND algorithm~\cite{rnd}.

\section{RELATED WORK}
One approach to curiosity-driven exploration is known as Intrinsically Motivated Goal Exploration Processes (IMGEPs) \cite{IMGEP}. IMGEPs equip the agent with a goal space, where each point is a vector of (target) features of behavioural outcomes. During exploration, the agent samples goals in this goal space by maximizing empirical competence progress using multi-armed bandits, enabling a learning curriculum where goals are progressively explored from simple to more complex. One limitation of IMGEPs is the need to manually engineer goal space representations, which is difficult for high-dimensional observations such as images.

Closely related to our approach is Go-Explore~\cite{go-explore}. In Go-Explore, the agent is returned to promising states from its previous experience to explore further. The implementation of this idea in \cite{go-explore} has some  drawbacks. The simulator's internal state is reset in order to return to promising states, which is not feasible for real robots. Our approach, on the other hand, trains a goal-conditioned policy to reach desired states. Moreover, novelty is measured by counting state visitation, which requires compact state representations that are heavily dependant on prior knowledge of the domain.

In order to alleviate some of the shortcomings of Go-Explore and intrinsic motivation methods,  Guo and Brunskill~\cite{DBLP:journals/corr/abs-1906-07805} use HER alongside a one-step prediction model to calculate novelty. While their method addresses the non-static nature of the intrinsic reward and does not require a reset like Go-Explore, it does not take full advantage of the data throughout the training due to the fact that both the goal relabeling and novelty assignment are done statically: the goal and novelty value for each state is picked only once when they are inserted into the replay buffer. Moreover, one-step prediction models can be difficult to train on high-dimensional state spaces and are susceptible to random noise from the environment. Indeed, \cite{DBLP:journals/corr/abs-1906-07805} did not show results with image inputs, and to the best of our knowledge, our work demonstrates the first application of goal-conditioned policies and intrinsic motivation with images.

The idea of quickly reaching novel states for efficient exploration was shown to be effective in the tabular case with algorithms such as Explicit Explore or Exploit ($E^3$) \cite{kearns2002near}. $E^3$ divides the state space into two sets, known and unknown, based on how thoroughly a given state has been visited. When exploring, a policy for reaching unknown states quickly is used. In the tabular case, a simple visitation count can be used to measure state novelty; our work extends this idea to high-dimensional state spaces by using a generalized novelty measure and a goal-conditioned policy.

\section{RESULTS}

In this section we evaluate our algorithm on a robotic manipulation environment using various downstream tasks. We show that using our data collection method we can significantly improve learning performance in downstream tasks, as compared to other collection strategies. We investigate the following questions:
\begin{enumerate}
    \item Can we improve performance in downstream learning tasks by collecting more diversified data? 
    \item Can self-supervised data collection be useful for learning multiple downstream tasks using the same data?
    \item Does our goal-conditioned algorithm collect better data than vanilla RL exploration?
\end{enumerate}

\subsection{Illustrative domain}

\begin{figure}[t]
\captionsetup[subfigure]{justification=centering}
\centering

\begin{subfigure}[t]{.23\textwidth}
  \footnotesize
  \centering
  \includegraphics[width=\linewidth]{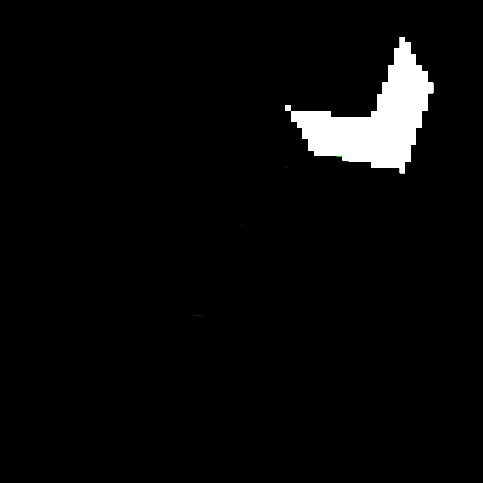}
\end{subfigure}\hspace{0.1cm}
\begin{subfigure}[t]{.23\textwidth}
  \footnotesize
  \centering
  \includegraphics[width=\linewidth]{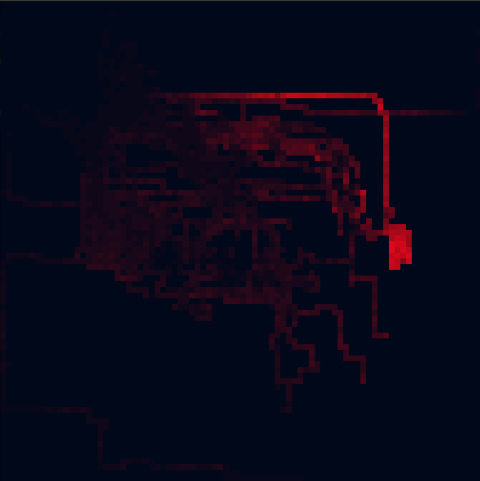}
\end{subfigure}

\caption{Illustrative experiment. On the left is the image state that the agent observes of the object moving against a black background. On the right is a visualization of the RND novelty values in states previously visited, the brighter the red the more novel the state is. The goal-conditioned agent will pick goals from the brightest region on the right side of the image.}
\label{fig:2D_env}
\vspace{-2em}
\end{figure}

We begin with a toy domain that will allow us to illustrate the benefits of our goal-conditioned exploration approach. The domain is motivated by a robotic problem of manipulating a rigid object using raw vision input. The observation is a white object against a black background, the object can move up, down, left and right by one pixel or rotate clock-wise and counter clock-wise with a resolution of 30 degrees. The size of the image is $84\times84$. The length of each episode is $250$ and the object is returned to the center at the beginning of each episode. An example can be seen in Figure~\ref{fig:2D_env}.

Our goal in this experiment is to test which approach covers as much of the state space as possible. To do so, we count the number of discrete states the agent visits during the training, where the state is the position and rotation of the object, resulting in a state space of size $84\times84\times12$.

\begin{figure}[t]
\captionsetup[subfigure]{justification=centering}
\centering

\begin{subfigure}[t]{.47\textwidth}
  \footnotesize
  \centering
  \includegraphics[width=\linewidth]{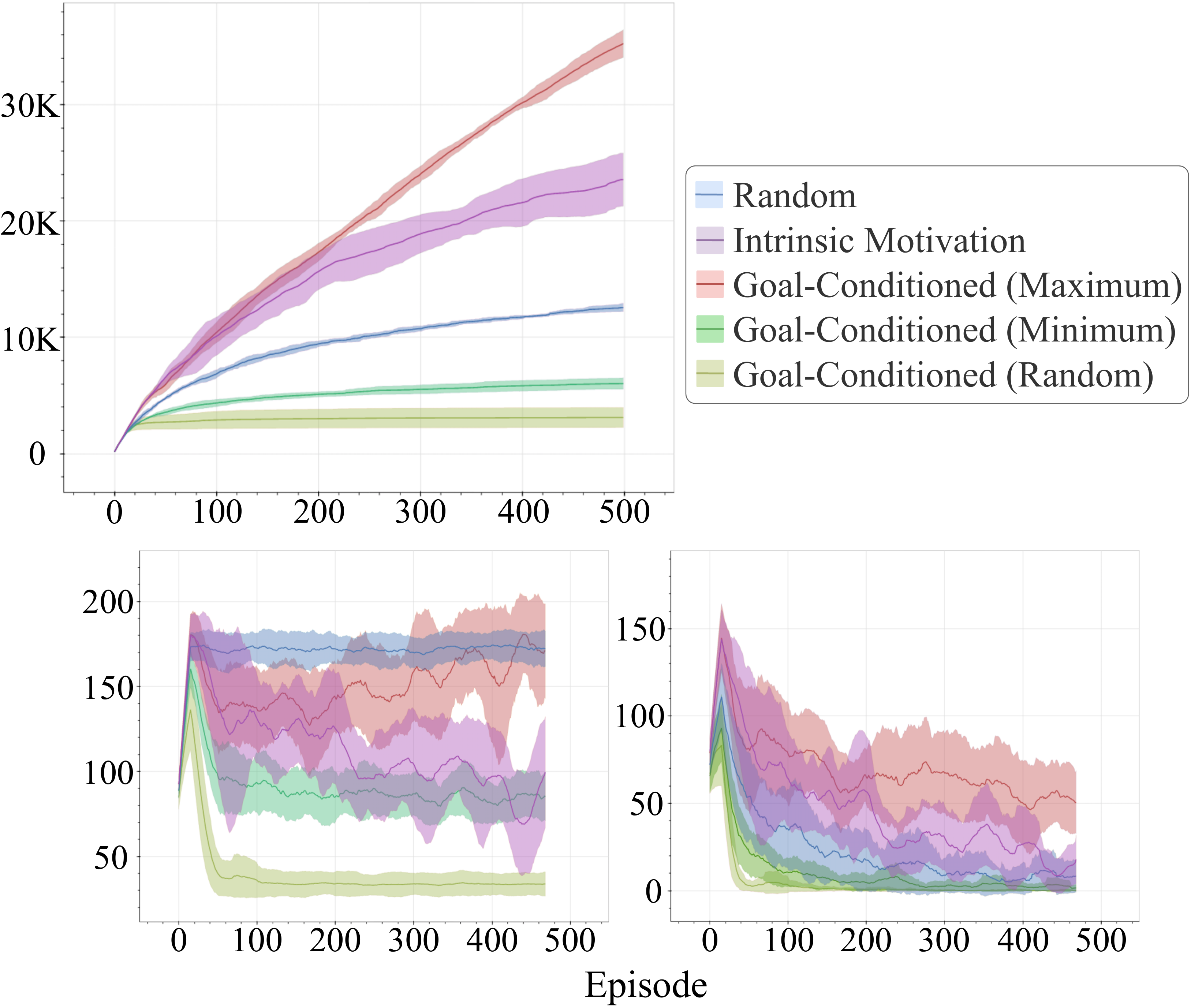}
\end{subfigure}

\caption{State visitation counts for the illustrative domain. On top we show the total number of  unique states the agent visited. On the bottom left we show the number of unique states the agent visited during each episode. Lastly, on the bottom right we show the number of unique states the agent visited during each episode which have also never been visited in previous episodes.}
\label{fig:2D_results_total}
\vspace{-1em}
\end{figure}

In addition to the baselines described above, we also consider an ablation where we change the goal picking strategy when acting (line 6 in Algorithm~\ref{alg1}) to either picking a random state from the replay buffer or a state with minimum novelty.
Figure~\ref{fig:2D_results_total} shows that RND novelty leads to better coverage of the state space compared to random exploration. Moreover, our algorithm exploits novelty information more efficiently than vanilla RL with intrinsic motivation. As expected, the goal-conditioned agent based on minimum novelty goals performs worst. Interestingly, the agent that picks random goals performs worse than the agent that behaves completely randomly, indicating that the goal picking strategy is an important component of the algorithm. We can also see that the random exploration reaches a plateau and doesn't keep exploring further, which means we won't achieve better exploration by simply collecting more samples with random actions.  

Additionally, we plot the average number of unique states visited within every episode, and the average number of unique and novel states visited every episode, i.e., the number of states that the agent visited for the first time over the entire training so far. Note that the goal-conditioned agent visits less unique states every episode than the random agent, but significantly more novel states. This is since the goal-conditioned approach leads the agent to discover \textit{novel regions}, all throughout the training process. 

\subsection{Simulated Robotic Manipulation Domains}
We next describe results on simulated robotic manipulation domains. We begin by describing our simulation setting and baseline comparisons, and then present our results.

\begin{figure}[t]
\captionsetup[subfigure]{justification=centering}
\centering
\begin{subfigure}[t]{.25\textwidth}
  \footnotesize
  \centering
  \includegraphics[width=\linewidth]{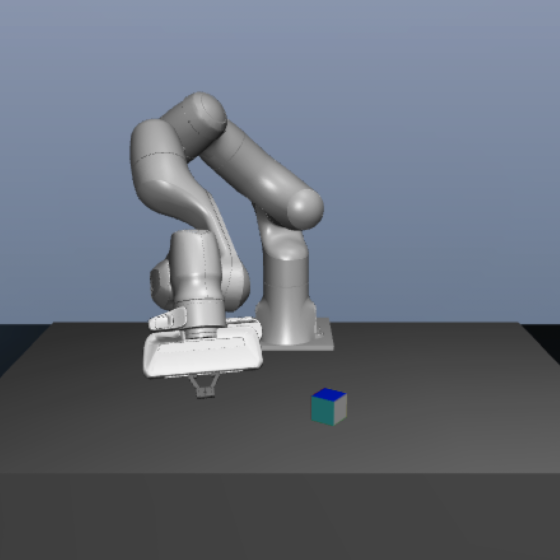}
\end{subfigure}\vspace{0.1cm}

\begin{subfigure}[t]{.09\textwidth}
  \centering
  \includegraphics[width=\linewidth]{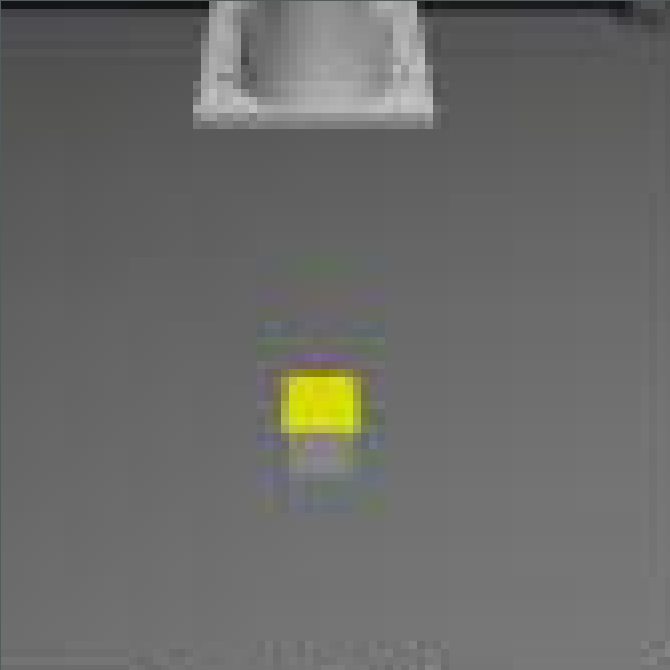}
  \caption*{Low \\ Novelty}
\end{subfigure}
\begin{subfigure}[t]{.09\textwidth}
  \centering
  \includegraphics[width=\linewidth]{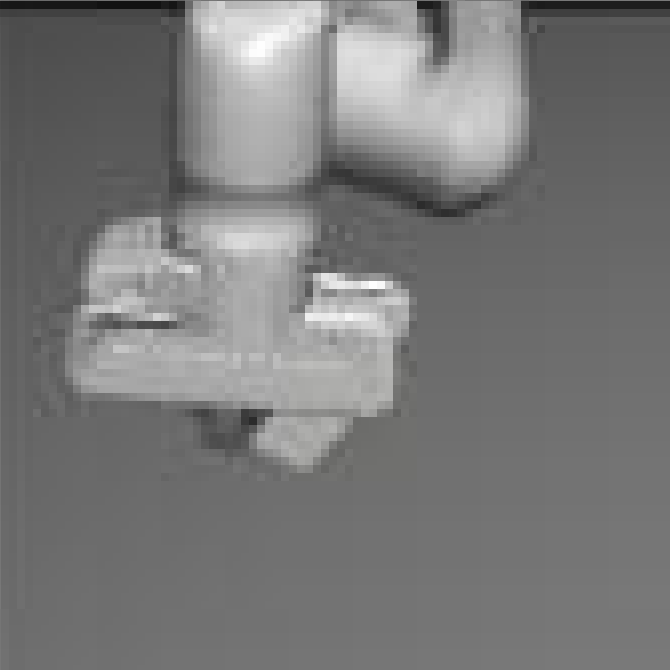}
\end{subfigure}
\begin{subfigure}[t]{.09\textwidth}
  \centering
  \includegraphics[width=\linewidth]{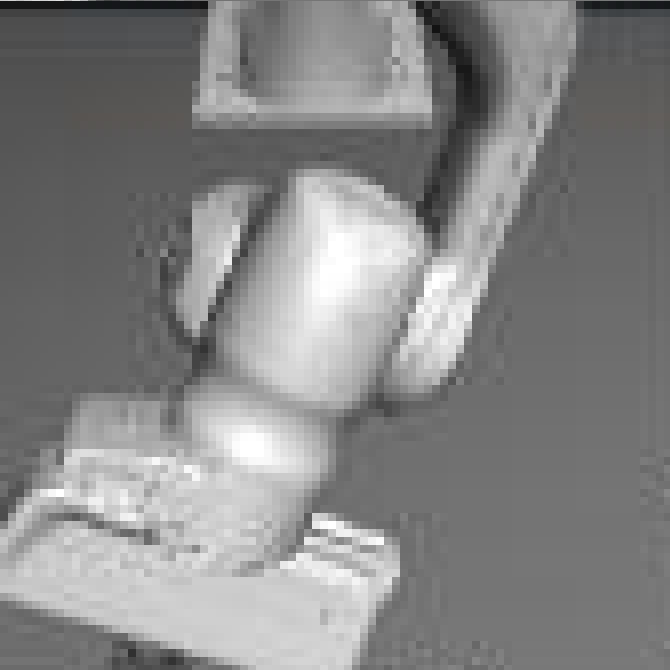}
\end{subfigure}
\begin{subfigure}[t]{.09\textwidth}
  \centering
  \includegraphics[width=\linewidth]{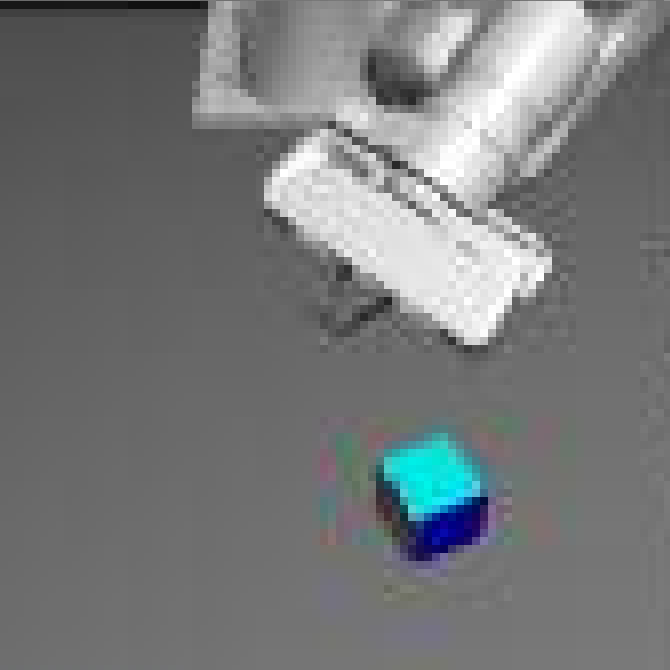}
\end{subfigure}
\begin{subfigure}[t]{.09\textwidth}
  \centering
  \includegraphics[width=\linewidth]{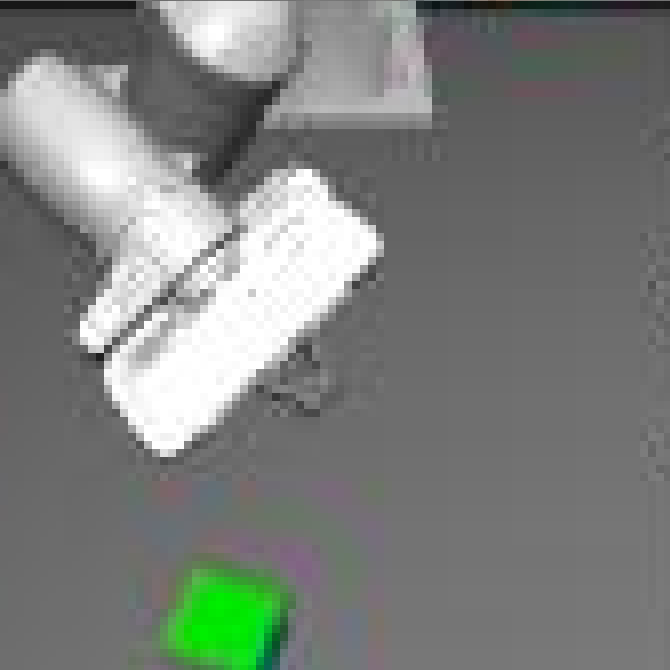}
  \caption*{High \\ Novelty}
\end{subfigure}

\caption{Simulation environment. Top: a simulated Franka Panda setup on a table and a rigid cube object. Bottom: examples of the observations that the RL agent and RND network get as input. The bottom row is ordered from left to right by the novelty of that state as given by the RND at the end of training.}
\label{fig:env}
\vspace{-1em}
\end{figure}

\textbf{Simulation environment}: We simulate a 7DoF Franka Panda robotic arm with a closed gripper and cartesian position control of the end-effector. The robot is positioned on a table, and a cube object with colored sides is placed in front of it. The robot can move freely in space, and can move and flip the cube on the table into different configurations. The camera observations are retrieved from a fixed camera positioned in front of the robot looking downwards at the table. An example of the setup can be seen in Figure \ref{fig:env}. For studying how the robot's presence in the image affects our exploration algorithms, we also consider an ablation where the robot is not rendered in the camera image. For the simulation we used Robosuite \cite{robosuite2020}.

\textbf{Baselines}: We compare our data collection method against two baselines: the first is a fixed random policy which samples actions uniformly from the action space of the robot. The second is a vanilla RND approach: we trained a TD3 agent that maximizes intrinsic reward calculated as novelty by RND. We emphasize that the same RND architecture and training schedule was used for the baseline and our method, with the only difference being how the RND novelty is used by the RL algorithm.

\textbf{Data collection and organization}: For data collection, we used the TD3 implementation in RL Coach \cite{coach}. An episode in our domain is set to $200$ transitions. For each method, we collect $5$ separate data-sets, where each data-set contains $M\cdot T=300K$ transitions (1500 episodes), and the RND network and TD3 agent are reset between collecting the data-sets. After each episode, the cube is reset to a fixed initial position with the yellow face up.

\textbf{Network architecture}: In this work we wish to disentangle the questions of training algorithm and architecture from exploration. Therefore, we chose popular neural network architectures, and used the same architectures for all methods we evaluated. Specifically, we used convolutional network architectures similar to \cite{dqn} for the agent and \cite{rnd} for the RND network. For the goal-conditioned agent, we condition the network on a goal image by concatenating it with the state image in the channel dimension. More details on specific parameters can be found in the appendix.

\subsubsection{Data-Set Collection Analysis}

\begin{figure}[t]
\centering
\begin{subfigure}{.47\textwidth}
  \centering
  \includegraphics[width=\linewidth]{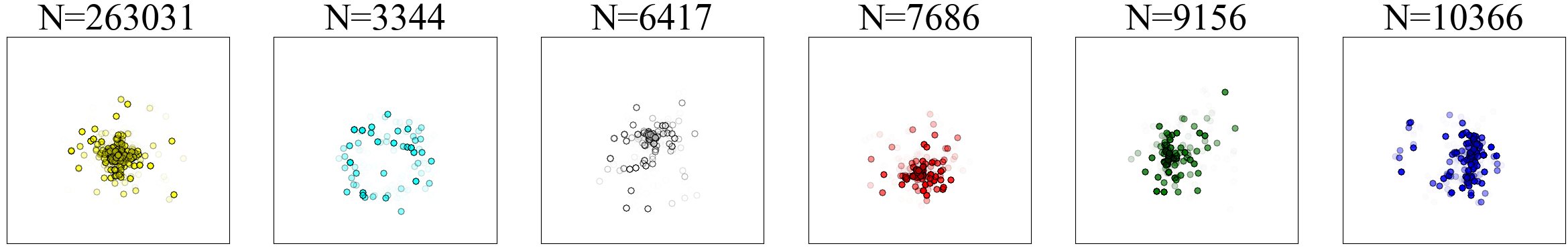}
  \caption{Random Agent}
\end{subfigure}
\begin{subfigure}{.47\textwidth}
  \centering
  \includegraphics[width=\linewidth]{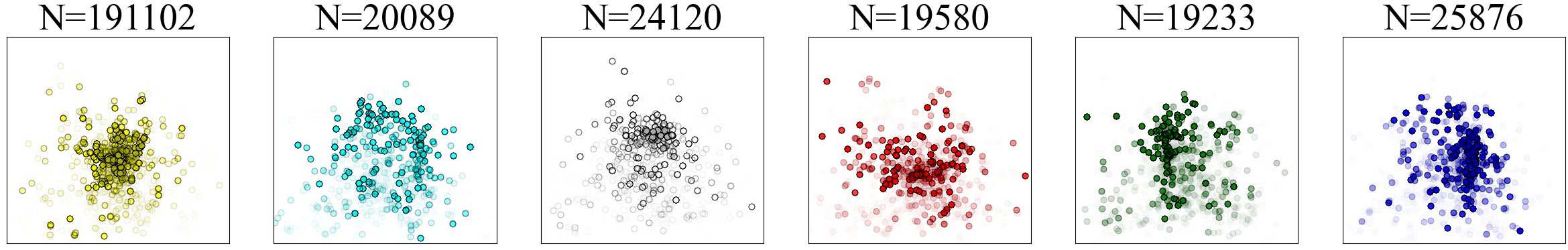}
  \caption{Intrinsic Reward Agent}
\end{subfigure}
\begin{subfigure}{.47\textwidth}
  \centering
  \includegraphics[width=\linewidth]{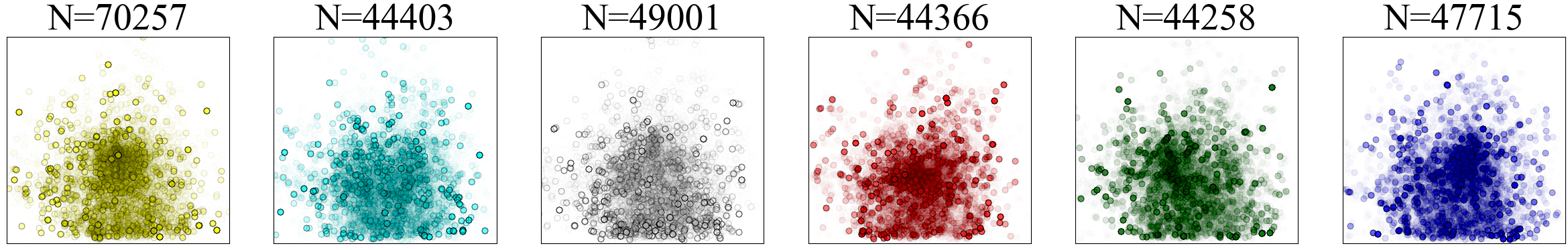}
  \caption{Goal-Conditioned Agent}
\end{subfigure}
\caption{Visualization of the data-sets collected using each method. The dots represent a position of the cube on the table as seen in the data-set, and the color corresponds to the color of the face at the top. The number at the top signifies that number of dots a plot contains for a certain color. Note that our goal-conditioned approach produces data that is significantly more diverse and evenly distributed than the baselines.}
\label{fig:scatter_color}
\vspace{-1em}
\end{figure}

As mentioned above, one of our objectives is to collect data that is diversified and evenly distributed. In Figure~\ref{fig:scatter_color} we visualize the collected data-sets by plotting the $x,y$ position of the cube on the table for every state in the data. We group the plots by the color of the upper face of the cube (recall that the cube is initialized with the yellow face up). The goal-conditioned exploration covers a significantly larger part of cube positions on the table, and does so more uniformly compared to baselines. Additionally, the distribution of top face color of the cube is more evenly distributed among the different colors, resulting in a less imbalanced data distribution, which is known to greatly effect learning in downstream tasks, as we demonstrate in our subsequent experiments.

It is important to note that even though the robot covers a significantly larger region of the input image than the cube (see Figure~\ref{fig:env}), our method was able to identify novelty in the cube position and orientation.

\subsubsection{Supervised Learning Tasks}

\begin{figure}[t]
    \centering
    \includegraphics[width=0.47\textwidth]{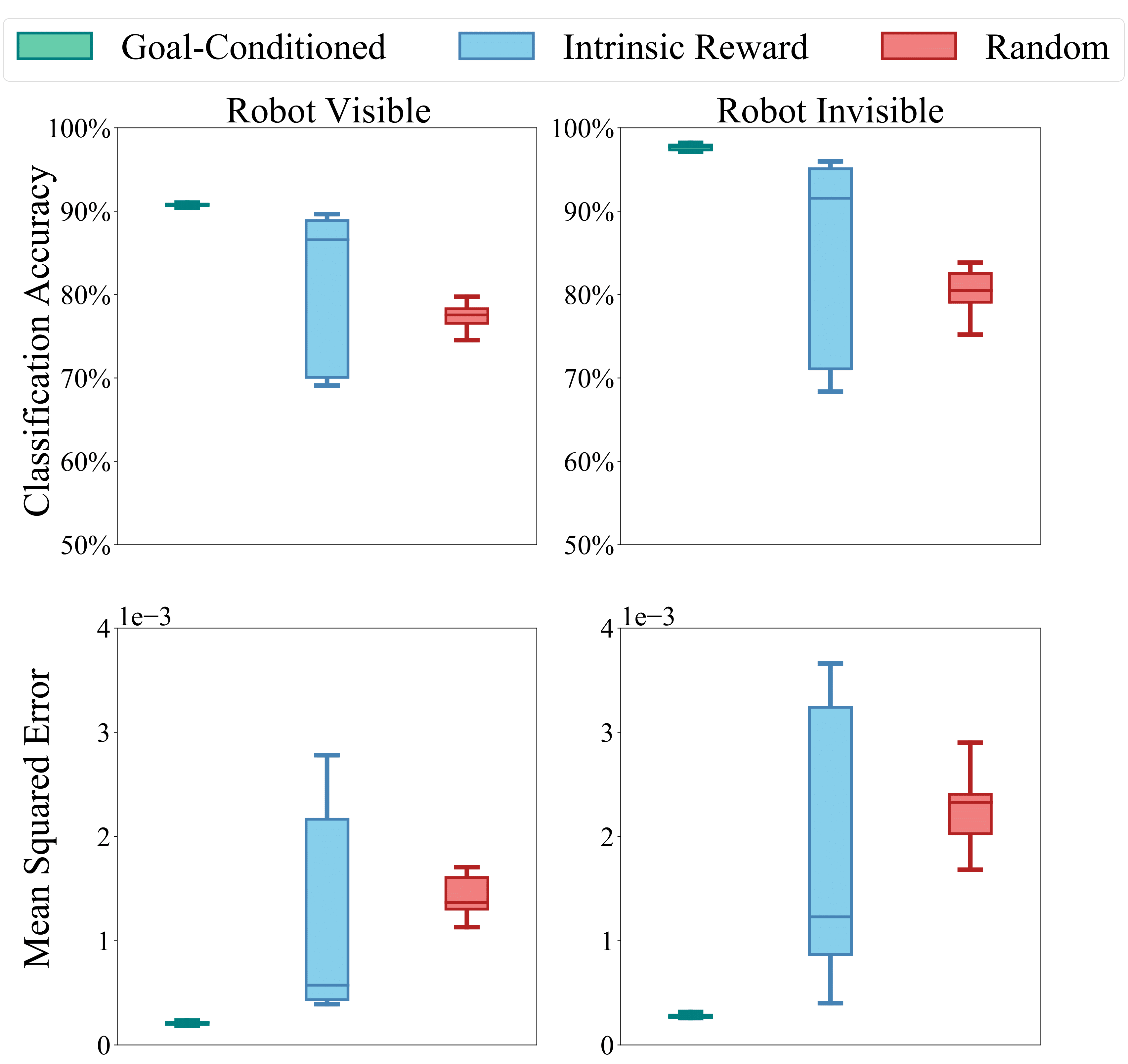}
    \caption{Supervised learning experiments. The upper row shows the classification accuracy of predicting the color of the upper face of the cube. The lower row shows the mean squared error of predicting the x, y, z position of the cube on the table.}
    \label{fig:supervised}
    \vspace{-1em}
\end{figure}

A common downstream task is supervised learning -- learn to predict some property of the scene such as the pose of an object. To evaluate the effect of the collected data on learning quality, we train a NN on two different prediction tasks, for which training labels can easily be extracted from the simulator:
\begin{itemize}
    \item Classification: predict the upper face color of the cube.
    \item Regression: predict the position of the cube on the table.
\end{itemize}

Since it's difficult to produce test data that is independent of all three methods, in order to keep our testing criteria as fair as possible, we produce a train/test split by randomly sampling $5\%$ of the examples from each data-set of each method to create a single test set for all the methods. The remaining $95\%$ samples constitute the training set. This procedure results in 5 training sets for each method and a single test set for all the methods.

The results are presented in Figure~\ref{fig:supervised}. For both classification and regression, the goal-conditioned exploration data-sets produced a higher median accuracy and lower MSE with little variance. Also note that the ablation of removing the robot from the image had a positive yet minor effect, showing the robustness of our approach.

\subsubsection{Offline Reinforcement Learning Tasks}\label{sss:offline}

We next evaluate our data collection method for training offline (a.k.a. Batch) RL agents. Specifically, we evaluate the performance of training a policy using Batch Constrained Q-Learning (BCQ\footnote{The BCQ implementation we used can be found \href{https://github.com/sfujim/BCQ/tree/master/continuous_BCQ}{\emph{here}}.}) \cite{DBLP:journals/corr/abs-1812-02900}, with the data-sets collected using different collection policies.

We consider three different tasks of varying difficulty:
\begin{itemize}
    \item Flip the cube to have a specific color facing up. This task is specified by a binary reward of $1$ when the correct color is on the upper face and $0$ otherwise. 
    \item Push the cube to the rightmost region of the table. The reward is proportional to horizontal distance between the goal region and the cube, specifically, $r=1-tanh(\rho(cube_y-\mathcal{T}_y))$, where $\rho$ is a scaling constant, and $\mathcal{T}_y$ is the threshold in the y direction of the region we want to push the cube into.
    \item Push the cube to the rightmost region of the table and flip it to a specific color. The reward is the sum of the two previous reward functions.
\end{itemize}

We emphasize that the rewards are not related to the behaviour of the robot during the data collection. In reality, defining rewards can be difficult. To focus our evaluation on data quality, however, we chose tasks for which rewards are easily constructed using data extracted from the simulator. 

\begin{figure}[t]
    \centering
    \includegraphics[width=0.46\textwidth]{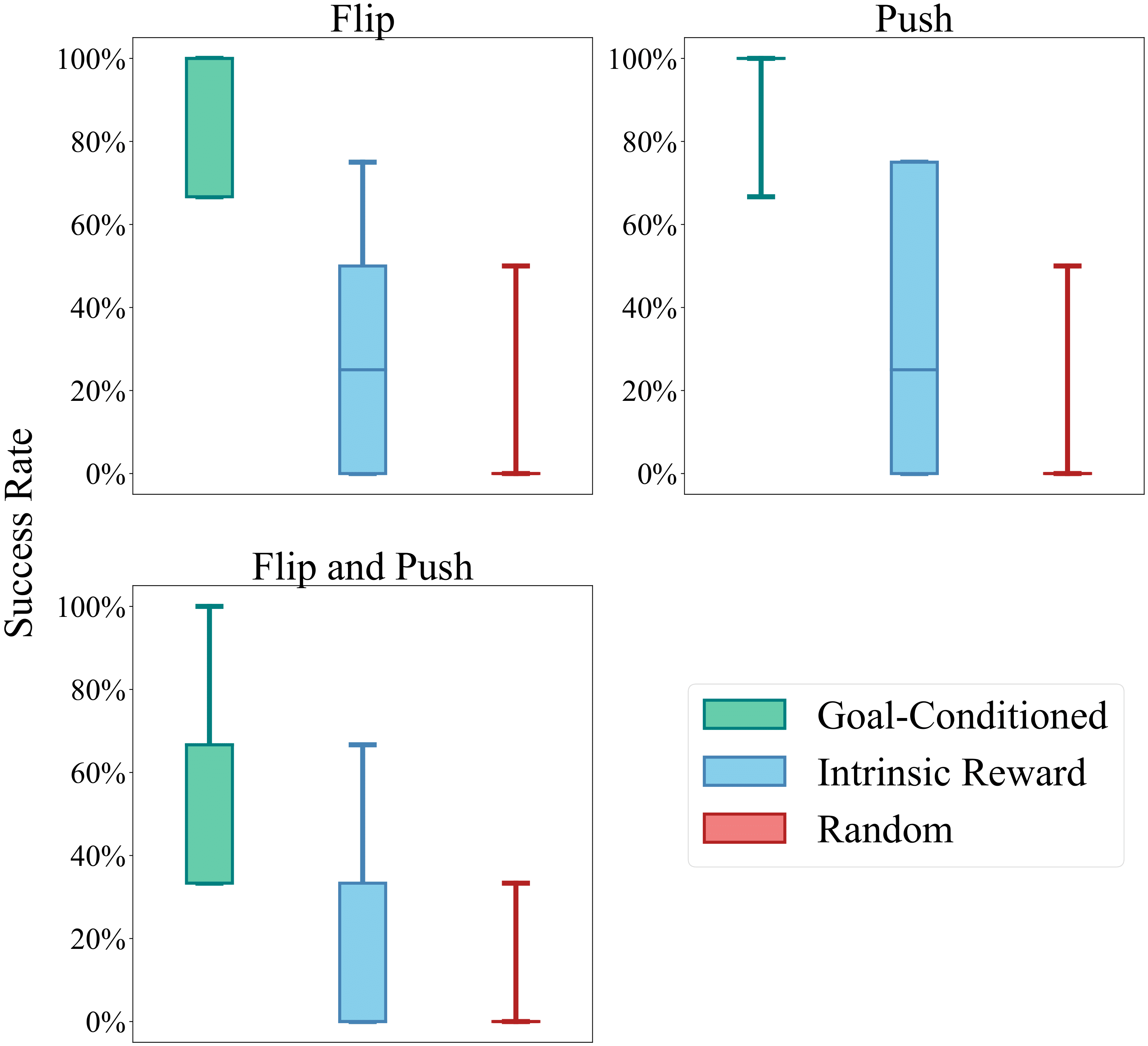}
    \caption{Offline learning experiments for the different tasks of pushing, flipping and their combination. The numbers indicate the percentage of the times the learned policy was successful at performing the task. For the pushing tasks, success indicates that the cube is in the target region.}
    \label{fig:offline}
\end{figure}

The results of the offline learning can be seen in Figure~\ref{fig:offline}. 
Similarly to the supervised learning experiments, the goal-conditioned exploration data-sets produced a higher success rate with lower variance. We also performed an ablation of hiding the robot in the images, which gave similar results; we omit it due to space constraints.

\section{CONCLUSIONS}
We proposed a simple-yet-effective method for generating diverse data-sets for offline robot learning. Our approach combines intrinsic motivation with goal-conditioned RL, to produce an agent that can quickly cover novel states.

In simulated robotic manipulation domains, our method produced significantly more diverse data-sets than common baseline methods, and led to improved performance on various downstream tasks. In future work we will evaluate our approach on real robot data collection, and on combinations of simulated and real data using sim-to-real methods.

\addtolength{\textheight}{-4cm}  



\section*{APPENDIX} \label{sec:appendix}

For all the algorithms we use the default hyper-parameters, with exception to those detailed in Table~\ref{appendix_table}. Additional details and implementation can be found \href{https://sites.google.com/view/efficientself-superviseddataco}{\emph{here}}.

\begin{table}[h]
\caption{Parameter Changes}
\label{appendix_table}
\begin{center}
\begin{tabular}{|c||c||c|}
\hline
Symbol & Usage & Value\\
\hline
\hline
$N$ & training cycles & 150\\
$M$ & \# episodes each training cycle & 10\\
 & probability of sampling self goal & 0.04\\
$K$ & RND training epochs & 4\\
 & TD3 actor architecture & DQN + Dense[300, 200]\\
  & TD3 critic architecture & DQN + Dense[400, 300]\\
    & TD3 learning rate & 0.0001\\
    & TD3 policy additive noise & linear schedule 1.5$\rightarrow$0.5\\
\hline
\hline
 & supervised learning architecture & DQN\\
 & supervised learning batch size & 128\\
 & supervised learning \# epochs & 25\\
 & supervised learning optimizer & ADAM(0.0001)\\
 & supervised learning \# seeds & 5\\
\hline
 & BCQ \# training steps & 50K\\
 & BCQ \# seeds & 3\\
$rho$ & BCQ \# push reward scaling & 10.0\\
$\mathcal{T}_y$ & BCQ \# push reward threshold & 18cm\\
\hline

\end{tabular}
\end{center}
\end{table}

\section*{ACKNOWLEDGMENT}
Aviv Tamar is partly funded by the Israel Science Foundation (ISF-759/19) and a grant from Intel Corporation.


\bibliographystyle{IEEEtran}
\bibliography{IEEEabrv, references}

\end{document}